\documentclass[runningheads]{llncs}
\usepackage{hyperref}
\usepackage[capitalize]{cleveref}
    \crefname{section}{Sec.}{Secs.}
    \Crefname{section}{Section}{Sections}
    \Crefname{table}{Table}{Tables}
    \crefname{table}{Tab.}{Tabs.}

\usepackage{makecell}
\usepackage{graphicx}

\usepackage{cuted}
\usepackage{amssymb}

\newcolumntype{C}[1]{>{\centering\arraybackslash}m{#1}}
\usepackage[T1]{fontenc}
\usepackage{graphicx}
\usepackage{orcidlink}

\begin{document}
\title{NVS-HO: A Benchmark for Novel View Synthesis of Handheld Objects}
%
%
\author{Musawar Ali\inst{1}\orcidlink{0000-0001-8684-3209} \and
Manuel Carranza-García\inst{2}\orcidlink{0000-0002-4729-8604} \and
Nicola Fioraio\inst{3}\orcidlink{0000-0001-9969-0555} \and
Samuele Salti\inst{1}\orcidlink{0000-0001-5609-426X} \and
Luigi Di Stefano\inst{1}\orcidlink{0000-0001-6014-6421}
}
\authorrunning{M. Ali et al.}
%
\institute{University of Bologna, Bologna, Italy \and
University of Sevilla, Sevilla, Spain \\
 \and
Eyecan.ai, Bologna, Italy \url{https://www.eyecan.ai/}\\
\email{\{musawar.ali2,samuele.salti, luigi.distefano\}@unibo.it}, mcarranzag@us.es, nicola.fioraio@eyecan.ai}
\maketitle              
\begin{abstract}
We propose NVS-HO, the first benchmark designed for novel view synthesis of handheld objects in real-world environments using only RGB inputs. 
Each object is recorded in two complementary RGB sequences: (1) a handheld sequence, where the object is manipulated in front of a static camera, and (2) a board sequence, where the object is fixed on a ChArUco board to provide accurate camera poses via marker detection. The goal of NVS-HO is to learn a NVS model that captures the full appearance of an object from (1), whereas (2) provides the ground-truth images used for evaluation.
To establish baselines, we consider both a classical SfM pipeline and a state-of-the-art pre-trained feed-forward neural network (VGGT) as pose estimators, and train NVS models based on NeRF and Gaussian Splatting.  Our experiments reveal significant performance gaps in current methods under unconstrained handheld conditions, highlighting the need for more robust approaches. NVS-HO thus offers a challenging real-world benchmark to drive progress in RGB-based novel view synthesis of handheld objects.

\end{abstract}
\begin{figure}[t] \centering \includegraphics[width=0.5\textwidth]{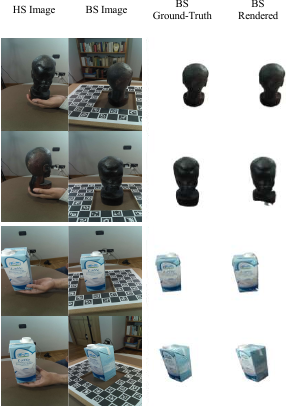} \caption{\textbf{ Two exemplar objects from our dataset.} Our dataset addresses novel view synthesis from images of handheld objects.
The columns show: (1) images from the HS, (2) images from the BS, (3) masked ground-truth  images and (4) masked rendered images.}
\label{fig:Main} 
\end{figure}
\section{Introduction\label{sec:Introduction}}

Hand-object interaction has received considerable attention in computer vision ~\cite{oikonomidis2018hands18}. Systems
capable of synthesizing realistic novel views from images of handheld
objects enable a wide range of applications in robotics,
augmented and virtual reality, digital content creation, and 3D
reconstruction~\cite{jiang2024hand,wu2024reconstructing,prakash2023learning}. Achieving high-quality novel view synthesis (NVS) in handheld settings requires accurate reasoning about object geometry and motion, often through intermediate 6-DoF pose estimation. While dedicated datasets have advanced handheld object pose estimation, the problem of NVS for handheld objects remains largely unexplored.



Capturing handheld objects where the object is moved while the camera remains fixed offers a particularly practical and accessible setup. This approach avoids camera drift, ensures stable framing, and, further guarantees complete 360\textdegree{} coverage with minimal effort. Compared to a moving camera, this setup is easier to achieve with simple hardware
and better reflects natural human interactions with objects, hence making it a strong foundation for benchmarking real-world NVS.

A wide range of datasets have been proposed for modeling and analyzing 3D hand-object interactions. Object-centric benchmarks such as LineMOD~\cite{hinterstoisser2012model}, YCB-Video~\cite{xiang2017posecnn}, and T-LESS~\cite{hodan2017t} paved the way  for modern 6-DoF pose estimation thus providing RGB-D images of texture-less objects together with accurate ground-truth poses derived from 3D models. Hand–object interaction datasets such as HO-3D~\cite{hampali2020honnotate} extend this direction by offering large-scale, multi-view data annotated with 3D hand pose and shape information. Other acquisition methods employ specialized equipment attached to the hand, including infrared markers~\cite{hillebrand2006inverse}, colour-coded gloves~\cite{wang2009real}, or electrical sensing devices~\cite{zimmerman1986hand} to facilitate pose estimation. While pose estimation has been enhanced by these datasets, they were not designed for novel view synthesis and therefore lack the ground-truth  views and evaluation protocols necessary to benchmark NVS methods. Furthermore, the techniques developed for existing datasets cannot estimate poses from RGB images alone, due to their reliance on depth information, synthetic data, or 3D object models.


In particular, NVS-oriented datasets have made strides towards capturing objects in real-world environments. Two of these benchmarks, CO3D~\cite{reizenstein:2021} and WildRGB-D~\cite{xia:2024} are object focused. The first provides large-scale, in-the-wild captures across multiple categories. The second brings objects into realistic contexts with background variation, but relies on RGB-D hardware rather than RGB-only inputs.  Differently,  HO-NeRF~\cite{Qu:2023}  is designed to model hand-object interactions by jointly estimating their poses as well as enabling NVS. However, lacking the ground-truth poses of objects,  the NVS methods built on this benchmark rely only on the hands to jointly estimate the poses of the objects and hands, so as to render \emph{both} the object and hands. In contrast, our benchmark provides ground-truth poses for objects and it is explicitly focused on wholly rendering the appearance of objects while getting rid of the partial occlusions induced by hands. 




Motivated by these limitations, we propose the first benchmark for NVS of handheld objects in real-world conditions using only RGB images, without relying on MANO~\cite{romero2022embodied} hand models or pre-scanned 3D object models. The dataset associated with our benchmark  
comprises two image sequences per object, as shown in~\cref{fig:Main}:
\begin{itemize}
    \item \textbf{Handheld Sequence (HS)} – The object is freely manipulated by hand while the camera remains static, realizing natural human interaction. This sequence is used to train a NVS model that can capture the full appearance of the object alone. 
    \item \textbf{Board Sequence (BS)} – The object is placed on a ChArUco board while the camera is moved across a set of capture positions. The markers present on the board enable accurate pose annotations, referred to as ground-truth poses. This sequence is used to evaluate the trained NVS model.
\end{itemize}

To establish baselines, we use COLMAP~\cite{schoenberger2016sfm}, a state-of-the-art structure-from-motion (SfM) pipeline, to recover camera poses for the HS. We also adapt VGGT~\cite{wang2025vggt}, a recent deep learning–based approach that leverages attention mechanisms to directly predict camera poses. Using the poses estimated by both methods, we train novel view synthesis (NVS) models, including Neural Radiance Fields (NeRFs)~\cite{mildenhall2021nerf} and Gaussian Splatting~\cite{kerbl20233d}, on the HS of each of the object captured in our dataset.  

To evaluate the considered methods, we render novel views from the same poses as the ground-truth ones and compare the rendered images to the corresponding BS images by standard metrics such as PSNR, SSIM, and LPIPS. As shown in ~\cref{fig:Main}, the comparison is carried out between masked image pairs where the background is represented  with the same colour (e.g. white). 
The key contributions of our paper can be summarized as follows. 
\begin{itemize}
    \item We introduce the first benchmark specifically designed for NVS of handheld objects in real-world environments using only RGB images. 
   \item We release a new dataset featuring RGB sequences that captures 67 objects.  Each object is recorded in two sequences: a HS featuring natural hand-object interaction and used for training,  a BS equipped with ground-truth poses designed for evaluation.  
    \item We propose a method-agnostic evaluation protocol to compare masked rendered images and masked ground-truth images. Key to this protocol is the ability to bring  the ground-truth viewpoints into the own 3D reference frame that may be adopted by any NVS method. 
    \item We benchmark state-of-the-art pipelines and reveal significant performance gaps under unconstrained handheld conditions, identifying  directions for future research in NVS.
\end{itemize}

\section{Related Work\label{sec:intro}}

This section reviews prior work on existing datasets including single-hand object datasets and  two-hand datasets.  Most existing hand object interaction datasets rely on MANO-based~\cite{romero2022embodied} hand models, pre-scanned 3D object models, or depth sensors to obtain ground-truth annotations, which limits their realism and scalability.
In contrast, our work introduces the first dataset that provides accurate object pose annotations without requiring complex hand or 3D object models. 
We also review the popular  NVS methodologies considered in our evaluation,

\subsection{Single-Hand objects}
Recently, several hand-object datasets have been introduced for grasp classification and action recognition tasks~\cite{rogez2015understanding,bambach2015lending,bullock2015yale,cai2015scalable,fathi2011learning}. However, despite the importance of hand action recognition, these datasets have rarely been used for NVS or object reconstruction, primarily due to the lack of ground-truth 6-DoF pose annotations.

Obtaining ground-truth pose information for hand-object interactions is a challenging task due to frequent hand occlusions and the typically low-texture appearance of objects. Moreover, in static scenes, it becomes nearly impossible to annotate object poses when the object is rotated around its own axis. These challenges also make it difficult to reliably integrate visual markers for pose annotation.

\subsection{Two-hand 3D pose datasets}

Annotating 3D hand poses becomes especially challenging when two hands interact or occlude each other. The Tzionas dataset~\cite{tzionas2016capturing} addressed this by combining RGB-D recordings with a realistic hand model that simulates skin movement~\cite{lewis2023pose}, and by training a model to recognize fingertips from carefully labeled examples. Similarly,~\cite{wang2020rgb2hands} introduced a dataset of two interacting hands captured with an RGB-D camera, using a depth-based tracker~\cite{mueller2019real} for annotation. Because the tracker was imperfect, they augmented the dataset with synthetic labels to improve accuracy. In a similar direction, ContactPose~\cite{brahmbhatt2020contactpose} relied on a richer setup of seven RGB cameras, three RGB-D sensors, and a thermal camera to capture detailed contact interactions between hands and objects.
Again, none of these dataset provides ground-truths and metrics to benchmark NVS methods.


\subsection{Novel view synthesis}

Novel view synthesis (NVS) aims to generate unseen viewpoints of a scene or object from a limited set of input images, and has become a central problem in 3D vision. Neural Radiance Fields (NeRFs)~\cite{mildenhall2021nerf} established the state of the art for NVS, while 3D Gaussian Splatting~\cite{kerbl20233d} introduced a novel rasterization-based approach. Both techniques achieve high-quality results, but their performance heavily depends on estimating camera poses accurately.  

More recently, Vision Transformer (ViT)-based approaches~\cite{dosovitskiy2020image} have explored fully data-driven, feed-forward  NVS, where explicit camera poses are not required. In particular, the Large View Synthesis Model (LVSM)~\cite{jin2024lvsm} represents a significant step forward in pose-free NVS. 
Similarly, VGGT~\cite{wang2025vggt} extends this paradigm by leveraging four input views and representing target viewpoints using Plücker rays. 
Nevertheless, these transformer-based approaches present limitations when scaling to large input sets (e.g., 100+ images). 
Thus, we do not include  feed-forward NVS models~\cite{jin2024lvsm,wang2025vggt}  among the baselines considered in our benchmark, leaving the investigation on whether they could be leveraged with unconstrained  handheld sequences as an interesting direction for future research.

\begin{figure}[t]
    \centering
    \includegraphics[width=1.0\textwidth]{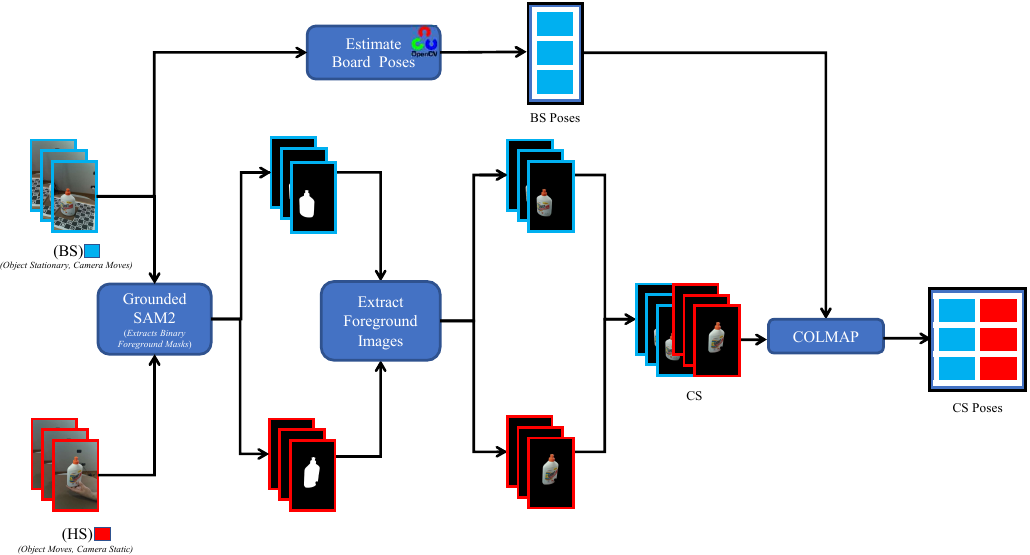}
    \caption{
    \textbf{Data processing pipeline.} For each object we record two monocular RGB sequences: an HS where the object is manipulated by a human while the camera is static, and a BS where the same object is fixed on a ChArUco board. BS poses are estimated with OpenCV via marker detection. Then,  Grounded SAM2~\cite{ren2024grounded} produces segmentation masks for the BS and HS, enabling foreground extraction. HS and BS are combined to produce CS. The CS combined with the fixed BS poses and the calibrated intrinsics are then fed into COLMAP that produces the resulting CS poses, where HS poses are estimated according to the units and coordinate system of the BS.}
    \label{fig:datasetcollection}
\end{figure}


\section{Dataset Creation}
\label{section:dataset_creation}

\noindent \textbf{Material preparation.}
We collected 67 common objects spanning two broad categories: grocery items and toys. The objects vary in size, shape, and texture, introducing diverse challenges for novel view synthesis. As already highlighted in the~\cref{sec:Introduction} and further detailed in this section,  two RGB sequences were acquired for each object, 
both taken  under controlled indoor lighting to ensure consistent image quality. 

\noindent \textbf{Camera.}
All sequences were captured with an OAK-D Lite camera\footnote{\href{https://docs.luxonis.com/hardware/products/OAK-D Lite}{https://docs.luxonis.com/hardware/products/OAK-D Lite}}, a compact RGB-D device with synchronized RGB and depth streams. In this work, though, we use only the RGB modality. 
For every recording we captured approximately 100–200 RGB frames at a resolution of $1080 \times 1920$ pixels, starting and ending from the same viewpoint to guarantee complete 360\textdegree{} coverage of the object.

\noindent \textbf{Calibration and Undistortion.}
The OAK-D Lite was calibrated using a standard chessboard pattern. 
The calibration images were captured at the same resolution as the dataset images.  
Calibration provides the intrinsic parameters matrix and the distortion coefficients. 
All RGB frames are undistorted using these parameters, and the camera model is converted to a pure pinhole representation.
  
\noindent \textbf{Image Acquisition.} For each object we record two complementary monocular RGB sequences: HS and BS (see~\cref{fig:Main}). 
The HS shows the object being freely manipulated by hand while the camera remains static. 
Although we capture image sequences rather than a continuous video, some frames from HS exhibit motion blur, which we keep to make the benchmark more realistic and challenging. These handheld sequences are used to estimate camera poses and train NVS methods.
The BS records the object fixed on a ChArUco board while the camera is moved around to capture a set of images that frame the object from different viewpoints. For these sequences, ground-truth poses are obtained using OpenCV~\cite{bradski2000opencv}. The detected 2D ChArUco corners are matched to their known 3D board coordinates. Then, \texttt{solvePnP} computes the per-frame camera rotation \(R_i\) and translation \(T_i\) relative to the board’s coordinate frame. These board poses provide accurate ground truth annotations for evaluation. 

\section{Benchmark}
\subsection{Dataset, Task, and Metrics}

\begin{figure}[t]
    \centering
    \includegraphics[width=\textwidth]{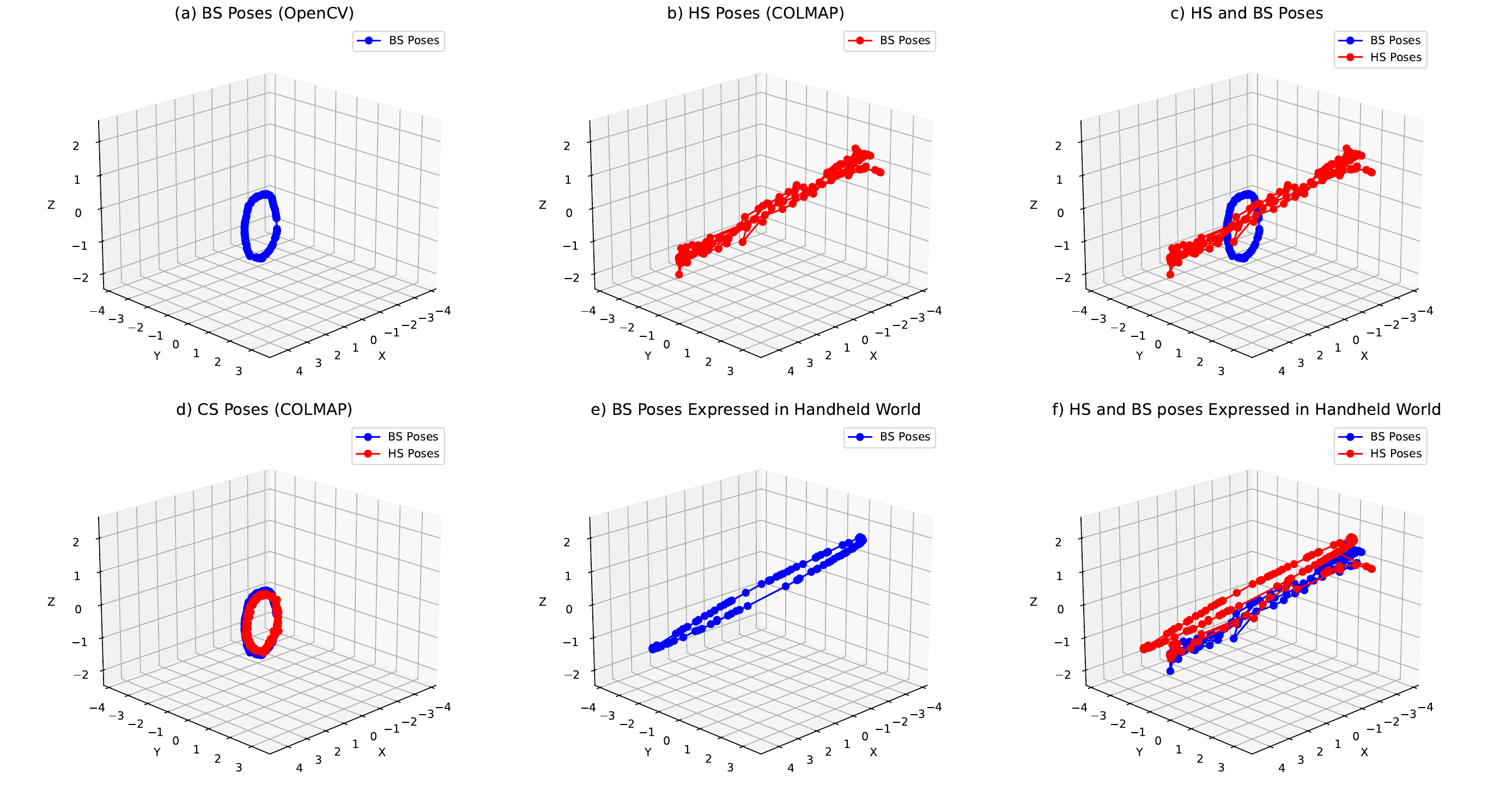}
    \caption{\textbf{Illustration of the pose alignment procedure.} (a) Camera poses estimated from the BS using a ChArUco board, expressed in metric scale with a world coordinate system defined by the calibration board. (b) Camera poses estimated from the HS using COLMAP, with an arbitrary scale and coordinate frame. (c) Overlay of (a) and (b), illustrating the misalignment due to differing coordinate systems and scales. (d) HS poses estimated by running COLMAP on the CS, with calibrated intrinsics and BS poses injected, ensuring both sequences share a common coordinate system. This trajectory serves as a reference frame bridging (a) and (b). (e) The BS poses from (a), transformed with a Sim(3) alignment computed via the Kabsch–Umeyama~\cite{umeyama2002least} algorithm, and now fully expressed in the HS coordinate frame. (f) Overlay of (b) and and the aligned BS poses from (e), confirming that the BS poses are now expressed in the HS coordinate system and units.
}
\label{fig:umeyama}
\end{figure}

    

\noindent \textbf{Dataset.} Using the procedures described in~\cref{section:dataset_creation}, we scanned 67 objects, capturing 100–200 views per object in both the HS and BS. The image sets densely cover 360\textdegree{} around each object, ensuring rich viewpoint diversity. 
The key data released in our benchmark consist of  undistorted RGB images from  both kinds of sequences, alongside the estimated camera intrinsics. 
In particular, for each object, the images from the HS are used to train NVS models, while those pertaining to the BS are used for evaluation purposes. The number of training and testing (i.e. evaluation)  images for each object is detailed in the Supplementary Material.

\noindent \textbf{Tasks.} Our benchmark focuses on evaluating novel view synthesis (NVS) of handheld objects in real-world conditions. Given a set of handheld images (i.e. the HS), a method must recover  6-DoF camera poses and use them to train a novel view synthesis model. 
Evaluation is carried out by querying the trained NVS model at viewpoints corresponding to the BS, where accurate ground-truth poses are available thanks to the detection of ChArUco markers.  The synthesized views are then compared to the corresponding masked RGB images from the BS frames. This setup assess jointly both the quality of novel view synthesis and the robustness of the underlying pose estimation.

\noindent \textbf{Pose alignment procedure.}
Since NVS methods rely on pose estimators that output trajectories in arbitrary scales and coordinate frames, we establish a procedure that can bring the test poses, defined in metric units within the coordinate frame of the BS, into the scale and coordinate frame estimated  from the HS by the considered NVS method. To this end, we start by constructing a Combined Sequence (CS) that acts as a bridge between the coordinate frame of the BS and that estimated by the NVS from the HS. 

As illustrated in~\cref{fig:datasetcollection}, the camera poses of the BS are first estimated in metric units (mm) and with respect to a world reference frame associated with the ChArUco board via standard marker detection. 
Then, we rely on Grounded SAM2 along with object-specific prompts~\cite{ren2024grounded} to segment the foreground object in the images of the BS and HS, thereby getting rid of the background in both, as well as of the person's hand in the latter. The foreground masks, thus, are used to set to black all background pixels in both sequences. The masked images belonging to both the BS and HS are merged into a single sequence, previously introduced as CS.  The next step of the procedure deals with running  COLMAP~\cite{schoenberger2016sfm} on the CS. In particular, we set the intrinsic parameters to the values estimated by camera calibration and inject as hard constraints the previously estimated poses of the BS. This forces COLMAP to recover the camera poses of the HS in the same metric units and world reference frame as those of the BS. Hence, the CS acts as a bridge between the two object captures and forms the foundation of our method-agnostic evaluation protocol. 




As anticipated, a NVS method feed with the HS will estimate camera poses in its own units and coordinate frame, and thus, to synthesize novel views, it will have to be queried with poses defined according to such units and frame. Hence, due to the poses of our test images being defined in the units and coordinate frame of the BS, to permit a quantitative evaluation based on the comparison between synthesized and test images, we need to bring the poses of the test images into the method's own units and reference frame. Purposely, 
we deploy the Kabsch–Umeyama algorithm~\cite{umeyama2002least} to estimate a similarity transformation (SIM(3)) based on the 3D-3D correspondences provided by the camera centers estimated by the NVS and those estimated by COLMAP via the CS. Since this transformation (a 3D roto-translation along with a scale change) brings 3D coordinates from the reference frame of the CS into the method's own one, it can be applied to the camera poses of the BS to define the viewpoints from which to synthesize the test images in order to evaluate the NVS method. 

An example reporting the different trajectories associated with the camera poses dealing with the BS, HS, CS and, finally, the aligned BS, is reported in  ~\cref{fig:umeyama}. 
In practice, given the camera poses estimated from the HS, the evaluation of a NVS method requires a simple, two-step, pose-alignment procedure: i) compute the SIM(3) transformation from the CS to the method's own reference frame and ii) apply the transformation to the test (i.e. BS poses). This alignment procedure makes it straightforward for users to transform the BS poses into their own estimated coordinate systems. Indeed, as part of our evaluation suite, we provide a script that, given the estimated HS poses, returns the aligned test poses. Finally, we point out that, along with the HS images, we also provide the associated segmentation masks obtained via Grounded SAM2~\cite{ren2024grounded}. Hence, users may choose to rely on our pipeline to mask out the background or develop their own one, or, perhaps, even to estimate poses on the original HS images,  without performing any explicit foreground segmentation step.


\noindent \textbf{Metrics.} Following standard practice in novel view synthesis, we evaluate the considered methods on our dataset using Peak Signal-to-Noise Ratio (PSNR)~\cite{huynh2008scope}, Structural Similarity Index Measure (SSIM)~\cite{wang2004image}, and Learned Perceptual Image Patch Similarity (LPIPS)~\cite{zhang2018unreasonable}. To correctly assess the rendering quality in our object-centric setting, we again leverage the segmentation masks associated with the BS. In particular, we define two complementary evaluation modes:

\begin{itemize}
    \item \textbf{Foreground Evaluation:} For PSNR, background pixels are masked out from the ground-truth images (i.e. those belonging to the BS) and the metric is computed solely within the object region. For SSIM and LPIPS, both the rendered and ground-truth images are cropped to tight bounding boxes defined by the foreground masks, focusing the evaluation on the object. Indeed, had the metrics been computed on whole images, the contribution of large background areas may have dominated the measured values, whereas the goal of our benchmark concerns the ability to accurately capture the full appearance of an object while it is being manipulated in front of a camera.
    
    \item \textbf{Background Evaluation:}  The foreground masks are inverted to select background regions, and all metrics (PSNR, SSIM, LPIPS) are computed exclusively on those areas. This is motivated by the fact that methods may produce artifacts in the background, which would not be penalized by the Foreground Evaluation. 
    To ensure fair comparison across methods, we normalize the background colour used in the evaluation.  Indeed, different NVS  approaches may render backgrounds via different colours (e.g. white, black or some gray), which skews the metrics  if a fixed background colour is chosen for the ground-truth images. To address this, we explicitly set the background colour of the ground-truth images to match that used by the specific NVS method under evaluation. This establishes a consistent evaluation protocol for future work. In particular, researchers willing to evaluate their method(s) will have to input the adopted background colour into an evaluation script provided with our benchmark, and the script  will modify the background colour of the ground-truth images accordingly. 
      We also point out that PSNR goes to infinity when comparing two identical images. This might occur in our benchmark because, as discussed above,  background pixels take the same colour in   rendered and ground-truth images. In particular, a method may render a background identical to the ground truth if pose estimation is nearly flawless or it fails to render any foreground at all, though in the latter circumstance the overall assessment will be penalized by the Foreground Evaluation. Due to the above eventualities, in the Background Evaluation we cap the PSNR to a very high value (i.e. 100) to enable score aggregation across images and objects.
\end{itemize}



The codebase, data, and evaluation scripts associated with our benchmark results will be publicly released upon acceptance of the paper. 


\subsection{Baseline Methods}
\label{sec:baseline_methods}

\noindent \textbf{Pose Estimation.} For pose estimation from the HS, we employ two state-of-the-art approaches: a classical SfM pipeline and a recent ViT-based model. 
To estimate poses by these methods we input masked images containing only the  foreground object extracted by Grounded SAM2,  as already discussed with reference to pose estimation from the CS. 

For SfM-based pose estimation, we adopt COLMAP~\cite{schoenberger2016sfm}. To estimate poses using COLMAP, all runs employ the sequential matcher, which is well-suited for video-style data. The sequential matcher relies on an overlap parameter, which determines how many neighboring frames are considered for feature matching. We tune this overlap individually for each object to account for variations in motion, frame density, and scene characteristics.

For ViT-based~\cite{dosovitskiy2020image} pose estimation, we adapt VGGT~\cite{wang2025vggt}, the current state-of-the-art in this field. Despite its general  effectiveness, we  found that VGGT struggles with sequences containing rather large uniformly coloured backgrounds,  which are common in our handheld-capture settings. To address this, we introduce several modifications to ensure that the model attends primarily to foreground object features. 
First, we compute the union of all foreground masks in the given HS  and crop images accordingly, updating the camera principal point after cropping. The cropped images are then processed by VGGT’s standard pipeline, including square padding, resizing, and feature extraction via DINO-V2~\cite{oquab2023dinov2}.
Next, we modify the input to the first VGGT attention block by pruning the tokens corresponding to background regions. This enforces attention on the object foreground, while substantially reducing memory usage. To ensure that every frame provides the same number of tokens to the attention layers, we identify the minimum number of foreground tokens present in the sequence and use this count consistently for all frames. For each image, we retain only the first tokens up to that count, while preserving their original positional encodings. These tokens are then passed through VGGT’s alternating attention layers. Finally, the pruned background tokens are reintroduced by scattering zeros into their original positions, ensuring compatibility with DPT~\cite{ranftl2021vision} heads.
For refinement, we leverage VGGT’s predicted point maps to perform bundle adjustment on the estimated poses, plugging in our calibrated (see~\cref{section:dataset_creation}), and later adjusted (see above) intrinsics in place of those estimated by the model.

\noindent \textbf{Novel View Synthesis.} For novel view synthesis, we establish the baselines considered in our benchmark by deploying  the Nerfstudio~\cite{tancik2023nerfstudio} library, using Nerfacto~\cite{tancik2023nerfstudio}, a method based on NeRF~\cite{mildenhall2021nerf}, and the Gaussian Splatting ~\cite{kerbl3Dgaussians} method Splatfacto. For both Nerfacto and Splatfacto we train models based on poses estimated by COLMAP and VGGT.  During training, we employ alpha transparency carving to better capture fine details and preserve clean object boundaries, which is critical for evaluating reconstruction quality on our object-centric dataset. In addition, we experiment with enabling pose optimization during training to assess its impact on performance.

The values of the key hyper-parameters concerning the considered pose estimation and novel view synthesis methods are reported  in the Supplementary Material.

\begin{figure}[t]
    \centering
    \includegraphics[width=\textwidth]{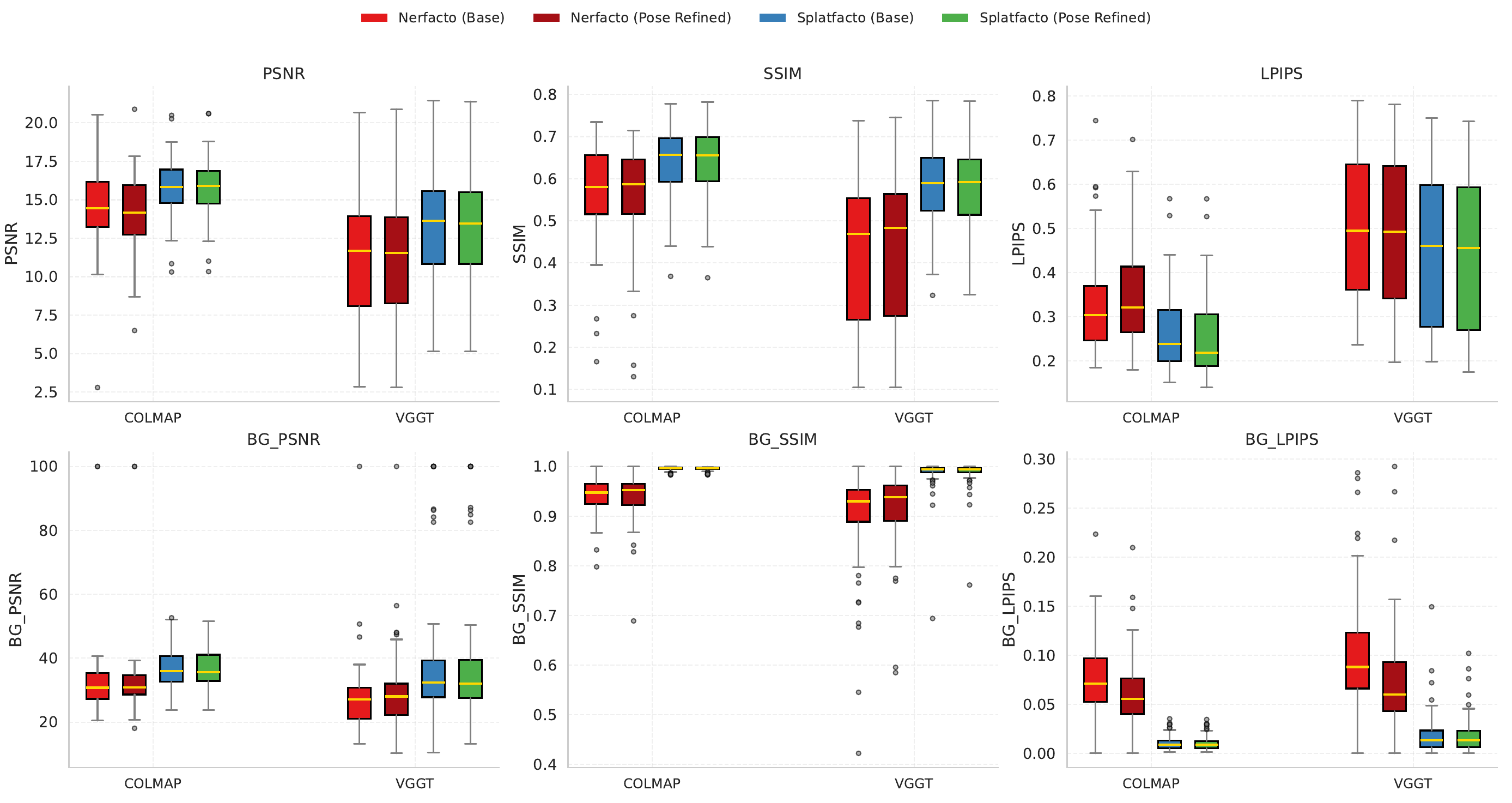}
    \caption{\textbf{NVS results.} 
Top and bottom rows report metrics dealing with Foreground and Background Evaluation, respectively. }
    \label{fig:results}
\end{figure}

\section{Benchmark Results}

\begin{figure}[t]
    \centering
    \includegraphics[width=0.75\textwidth]{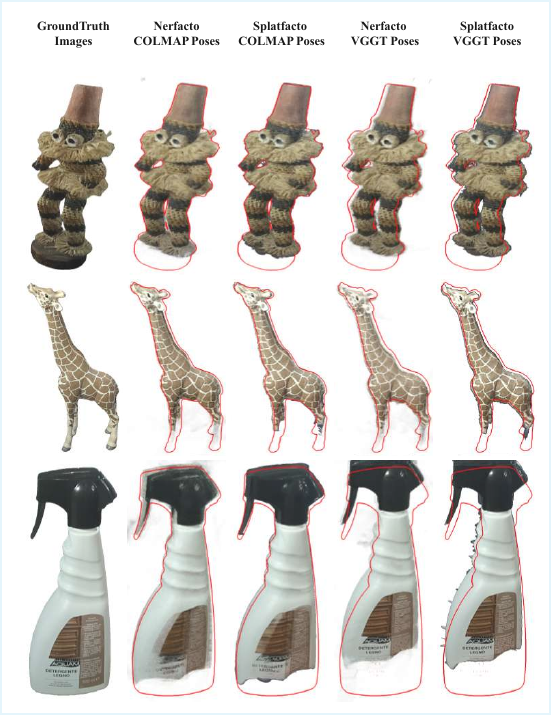}
    \caption{\textbf{Qualitative NVS results.} Red edges denote the ground-truth object masks. 
    .}
    \label{fig:qualitative comparison}
\end{figure}

In~\cref{fig:results}, we present results on our dataset NVS-HO, where camera poses were estimated using COLMAP~\cite{schoenberger2016sfm} and VGGT~\cite{wang2025vggt} and subsequently used to train Nerfacto~\cite{tancik2023nerfstudio} and Splatfacto~\cite{kerbl20233d}, both with and without pose optimization. We report the results as box plots to capture the distribution of the metrics across objects, due to the large number of objects and the large variance in the results. The same results are reported in tabular form in the supplementary material. The most evident result is that COLMAP outperforms VGGT across all metrics, exhibiting both better median values as well as less variance in the metrics. Hence, for reliable pose estimation in handheld scenarios, traditional SfM seems still preferable. Another interesting result is that Splatfacto consistently outperforms Nerfacto across all evaluation metrics. Yet, both the median PSNR and SSIM are low in absolute terms, e.g. the median PSNR of Splatfacto with COLMAP poses is about 16. Moreover, activating pose refinement at training time brings only marginal improvements. Since Nerfacto and Splatfacto usually deliver very good results in more controlled setups, where poses are known or can be estimated reliably, these findings suggest that more accurate pose initialization is required to improve metrics in our benchmark. 
This trend is also evident in the qualitative results displayed in~\cref{fig:qualitative comparison}, where we can see that, although  COLMAP yields renderings that align more tightly to ground-truth object boundaries compared to VGGT, both methods cannot fully render the object from the considered novel viewpoints. In the Supplementary Material, we provide detailed results across all objects.

\section{Final Discussion}

We have presented NVS-HO, the first benchmark dedicated to NVS of handheld objects in unconstrained real-world conditions using only RGB inputs. Unlike prior datasets, our benchmark does not rely on pre-scanned 3D object models or MANO hand models. 
Our experiments with state-of-the-art methods reveal that current NVS pipelines struggle under handheld scenarios, highlighting the challenges set forth by this very practical and accessible  acquisition setup, where the need to deal with hand-induced partial occlusions makes it more difficult learning the full appearance of the object. Our results underscore the need for more robust approaches that may tackle the peculiar difficulties posed by our benchmark. 
Hence, we believe that NVS-HO establishes a valuable foundation for the community, offering both a practical benchmark and a standardized evaluation protocol to drive future research in novel view synthesis of handheld objects.

%
%
%
 \bibliographystyle{splncs04}
 \bibliography{main.bib}

@String(CVPR= {IEEE Conf. Comput. Vis. Pattern Recog.})

@String(ICCV= {Int. Conf. Comput. Vis.})

@String(ECCV= {Eur. Conf. Comput. Vis.})

@String(TOG= {ACM Trans. Graph.})

@String(AAAI = {AAAI})

@String(CVPR  = {CVPR})

@String(ICCV  = {ICCV})

@String(ECCV  = {ECCV})

@String(TOG   = {ACM TOG})

@article{tzionas2016capturing,
  title={Capturing hands in action using discriminative salient points and physics simulation},
  author={Tzionas, Dimitrios and Ballan, Luca and Srikantha, Abhilash and Aponte, Pablo and Pollefeys, Marc and Gall, Juergen},
  journal={International Journal of Computer Vision},
  volume={118},
  pages={172--193},
  year={2016},
  publisher={Springer}
}

@inproceedings{lewis2023pose,
author = {Lewis, J. P. and Cordner, Matt and Fong, Nickson},
title = {Pose space deformation: a unified approach to shape interpolation and skeleton-driven deformation},
year = {2000},
isbn = {1581132085},
publisher = {ACM Press/Addison-Wesley Publishing Co.},
address = {USA},
url = {https://doi.org/10.1145/344779.344862},
doi = {10.1145/344779.344862},
booktitle = {Proceedings of the 27th Annual Conference on Computer Graphics and Interactive Techniques},
pages = {165–172},
numpages = {8},
series = {SIGGRAPH '00}
}

@article{wang2020rgb2hands,
  title={Rgb2hands: real-time tracking of 3d hand interactions from monocular rgb video},
  author={Wang, Jiayi and Mueller, Franziska and Bernard, Florian and Sorli, Suzanne and Sotnychenko, Oleksandr and Qian, Neng and Otaduy, Miguel A and Casas, Dan and Theobalt, Christian},
  journal={ACM Transactions on Graphics (ToG)},
  volume={39},
  number={6},
  pages={1--16},
  year={2020},
  publisher={ACM New York, NY, USA}
}

@inproceedings{fathi2011learning,
  title={Learning to recognize objects in egocentric activities},
  author={Fathi, Alireza and Ren, Xiaofeng and Rehg, James M},
  booktitle={CVPR 2011},
  pages={3281--3288},
  year={2011},
  organization={IEEE}
}

@inproceedings{cai2015scalable,
  title={A scalable approach for understanding the visual structures of hand grasps},
  author={Cai, Minjie and Kitani, Kris M and Sato, Yoichi},
  booktitle={2015 IEEE International Conference on Robotics and Automation (ICRA)},
  pages={1360--1366},
  year={2015},
  organization={IEEE}
}

@inproceedings{ranftl2021vision,
  title={Vision transformers for dense prediction},
  author={Ranftl, Ren{\'e} and Bochkovskiy, Alexey and Koltun, Vladlen},
  booktitle={Proceedings of the IEEE/CVF international conference on computer vision},
  pages={12179--12188},
  year={2021}
}

@article{oquab2023dinov2,
  title={Dinov2: Learning robust visual features without supervision},
  author={Oquab, Maxime and Darcet, Timoth{\'e}e and Moutakanni, Th{\'e}o and Vo, Huy and Szafraniec, Marc and Khalidov, Vasil and Fernandez, Pierre and Haziza, Daniel and Massa, Francisco and El-Nouby, Alaaeldin and others},
  journal={arXiv preprint arXiv:2304.07193},
  year={2023}
}

@inproceedings{wang2025vggt,
  title={Vggt: Visual geometry grounded transformer},
  author={Wang, Jianyuan and Chen, Minghao and Karaev, Nikita and Vedaldi, Andrea and Rupprecht, Christian and Novotny, David},
  booktitle={Proceedings of the Computer Vision and Pattern Recognition Conference},
  pages={5294--5306},
  year={2025}
}

@article{romero2022embodied,
  title={Embodied hands: Modeling and capturing hands and bodies together},
  author={Romero, Javier and Tzionas, Dimitrios and Black, Michael J},
  journal={arXiv preprint arXiv:2201.02610},
  year={2022}
}

@article{bullock2015yale,
  title={The Yale human grasping dataset: Grasp, object, and task data in household and machine shop environments},
  author={Bullock, Ian M and Feix, Thomas and Dollar, Aaron M},
  journal={The International Journal of Robotics Research},
  volume={34},
  number={3},
  pages={251--255},
  year={2015},
  publisher={SAGE Publications Sage UK: London, England}
}

@inproceedings{bambach2015lending,
  title={Lending a hand: Detecting hands and recognizing activities in complex egocentric interactions},
  author={Bambach, Sven and Lee, Stefan and Crandall, David J and Yu, Chen},
  booktitle={Proceedings of the IEEE international conference on computer vision},
  pages={1949--1957},
  year={2015}
}

@inproceedings{rogez2015understanding,
  title={Understanding everyday hands in action from rgb-d images},
  author={Rogez, Gr{\'e}gory and Supancic, James S and Ramanan, Deva},
  booktitle={Proceedings of the IEEE international conference on computer vision},
  pages={3889--3897},
  year={2015}
}

@inproceedings{schoenberger2016sfm,
    author={Sch\"{o}nberger, Johannes Lutz and Frahm, Jan-Michael},
    title={Structure-from-Motion Revisited},
    booktitle={Conference on Computer Vision and Pattern Recognition (CVPR)},
    year={2016},
}

@article{zimmerman1986hand,
  title={A hand gesture interface device},
  author={Zimmerman, Thomas G and Lanier, Jaron and Blanchard, Chuck and Bryson, Steve and Harvill, Young},
  journal={ACM Sigchi Bulletin},
  volume={18},
  number={4},
  pages={189--192},
  year={1986},
  publisher={ACM New York, NY, USA}
}

@article{wang2009real,
  title={Real-time hand-tracking with a color glove},
  author={Wang, Robert Y and Popovi{\'c}, Jovan},
  journal={ACM transactions on graphics (TOG)},
  volume={28},
  number={3},
  pages={1--8},
  year={2009},
  publisher={ACM New York, NY, USA}
}

@inproceedings{hillebrand2006inverse,
  title={Inverse kinematic infrared optical finger tracking},
  author={Hillebrand, Gerrit and Bauer, Martin and Achatz, Kurt and Klinker, Gudrun and Oferl, Am},
  booktitle={Proceedings of the 9th International Conference on Humans and Computers (HC 2006), Aizu, Japan},
  pages={6--9},
  year={2006}
}

@inproceedings{hinterstoisser2012model,
  title={Model based training, detection and pose estimation of texture-less 3d objects in heavily cluttered scenes},
  author={Hinterstoisser, Stefan and Lepetit, Vincent and Ilic, Slobodan and Holzer, Stefan and Bradski, Gary and Konolige, Kurt and Navab, Nassir},
  booktitle={Asian conference on computer vision},
  pages={548--562},
  year={2012},
  organization={Springer}
}

@inproceedings{hodan2017t,
  title={T-LESS: An RGB-D dataset for 6D pose estimation of texture-less objects},
  author={Hodan, Tom{\'a}{\v{s}} and Haluza, Pavel and Obdr{\v{z}}{\'a}lek, {\v{S}}tep{\'a}n and Matas, Jiri and Lourakis, Manolis and Zabulis, Xenophon},
  booktitle={2017 IEEE Winter Conference on Applications of Computer Vision (WACV)},
  pages={880--888},
  year={2017},
  organization={IEEE}
}

@article{bradski2000opencv,
  title={The opencv library.},
  author={Bradski, Gary},
  journal={Dr. Dobb's Journal: Software Tools for the Professional Programmer},
  volume={25},
  number={11},
  pages={120--123},
  year={2000},
  publisher={Miller Freeman Inc.}
}

@article{kerbl20233d,
  title={3d gaussian splatting for real-time radiance field rendering.},
  author={Kerbl, Bernhard and Kopanas, Georgios and Leimk{\"u}hler, Thomas and Drettakis, George},
  journal={ACM Trans. Graph.},
  volume={42},
  number={4},
  pages={139--1},
  year={2023}
}

@article{mildenhall2021nerf,
  title={Nerf: Representing scenes as neural radiance fields for view synthesis},
  author={Mildenhall, Ben and Srinivasan, Pratul P and Tancik, Matthew and Barron, Jonathan T and Ramamoorthi, Ravi and Ng, Ren},
  journal={Communications of the ACM},
  volume={65},
  number={1},
  pages={99--106},
  year={2021},
  publisher={ACM New York, NY, USA}
}

@article{dosovitskiy2020image,
  title={An image is worth 16x16 words: Transformers for image recognition at scale},
  author={Dosovitskiy, Alexey and Beyer, Lucas and Kolesnikov, Alexander and Weissenborn, Dirk and Zhai, Xiaohua and Unterthiner, Thomas and Dehghani, Mostafa and Minderer, Matthias and Heigold, Georg and Gelly, Sylvain and others},
  journal={arXiv preprint arXiv:2010.11929},
  year={2020}
}

@article{jin2024lvsm,
  title={Lvsm: A large view synthesis model with minimal 3d inductive bias},
  author={Jin, Haian and Jiang, Hanwen and Tan, Hao and Zhang, Kai and Bi, Sai and Zhang, Tianyuan and Luan, Fujun and Snavely, Noah and Xu, Zexiang},
  journal={arXiv preprint arXiv:2410.17242},
  year={2024}
}

@misc{ren2024grounded,
      title={Grounded SAM: Assembling Open-World Models for Diverse Visual Tasks}, 
      author={Tianhe Ren and Shilong Liu and Ailing Zeng and Jing Lin and Kunchang Li and He Cao and Jiayu Chen and Xinyu Huang and Yukang Chen and Feng Yan and Zhaoyang Zeng and Hao Zhang and Feng Li and Jie Yang and Hongyang Li and Qing Jiang and Lei Zhang},
      year={2024},
      eprint={2401.14159},
      archivePrefix={arXiv},
      primaryClass={cs.CV}
}

@inproceedings{zhang2018unreasonable,
  title={The unreasonable effectiveness of deep features as a perceptual metric},
  author={Zhang, Richard and Isola, Phillip and Efros, Alexei A and Shechtman, Eli and Wang, Oliver},
  booktitle={Proceedings of the IEEE conference on computer vision and pattern recognition},
  pages={586--595},
  year={2018}
}

@article{wang2004image,
  title={Image quality assessment: from error visibility to structural similarity},
  author={Wang, Zhou and Bovik, Alan C and Sheikh, Hamid R and Simoncelli, Eero P},
  journal={IEEE transactions on image processing},
  volume={13},
  number={4},
  pages={600--612},
  year={2004},
  publisher={IEEE}
}

@article{huynh2008scope,
  title={Scope of validity of PSNR in image/video quality assessment},
  author={Huynh-Thu, Quan and Ghanbari, Mohammed},
  journal={Electronics letters},
  volume={44},
  number={13},
  pages={800--801},
  year={2008},
  publisher={IET}
}

@article{umeyama2002least,
  title={Least-squares estimation of transformation parameters between two point patterns},
  author={Umeyama, Shinji},
  journal={IEEE Transactions on pattern analysis and machine intelligence},
  volume={13},
  number={4},
  pages={376--380},
  year={2002},
  publisher={IEEE}
}

@inproceedings{tancik2023nerfstudio,
  title={Nerfstudio: A modular framework for neural radiance field development},
  author={Tancik, Matthew and Weber, Ethan and Ng, Evonne and Li, Ruilong and Yi, Brent and Wang, Terrance and Kristoffersen, Alexander and Austin, Jake and Salahi, Kamyar and Ahuja, Abhik and others},
  booktitle={ACM SIGGRAPH 2023 conference proceedings},
  pages={1--12},
  year={2023}
}

@Article{kerbl3Dgaussians,
      author       = {Kerbl, Bernhard and Kopanas, Georgios and Leimk{\"u}hler, Thomas and Drettakis, George},
      title        = {3D Gaussian Splatting for Real-Time Radiance Field Rendering},
      journal      = {ACM Transactions on Graphics},
      number       = {4},
      volume       = {42},
      month        = {July},
      year         = {2023},
      url          = {https://repo-sam.inria.fr/fungraph/3d-gaussian-splatting/}
}

@article{xiang2017posecnn,
  title={Posecnn: A convolutional neural network for 6d object pose estimation in cluttered scenes},
  author={Xiang, Yu and Schmidt, Tanner and Narayanan, Venkatraman and Fox, Dieter},
  journal={arXiv preprint arXiv:1711.00199},
  year={2017}
}

@inproceedings{oikonomidis2018hands18,
  title={Hands18: Methods, techniques and applications for hand observation},
  author={Oikonomidis, Iason and Garcia-Hernando, Guillermo and Yao, Angela and Argyros, Antonis and Lepetit, Vincent and Kim, Tae-Kyun},
  booktitle={Proceedings of the European Conference on Computer Vision (ECCV) Workshops},
  pages={0--0},
  year={2018}
}

@inproceedings{hampali2020honnotate,
  title={Honnotate: A method for 3d annotation of hand and object poses},
  author={Hampali, Shreyas and Rad, Mahdi and Oberweger, Markus and Lepetit, Vincent},
  booktitle={Proceedings of the IEEE/CVF conference on computer vision and pattern recognition},
  pages={3196--3206},
  year={2020}
}

@inproceedings{brahmbhatt2020contactpose,
  title={ContactPose: A dataset of grasps with object contact and hand pose},
  author={Brahmbhatt, Samarth and Tang, Chengcheng and Twigg, Christopher D and Kemp, Charles C and Hays, James},
  booktitle={Computer Vision--ECCV 2020: 16th European Conference, Glasgow, UK, August 23--28, 2020, Proceedings, Part XIII 16},
  pages={361--378},
  year={2020},
  organization={Springer}
}

@article{mueller2019real,
  title={Real-time pose and shape reconstruction of two interacting hands with a single depth camera},
  author={Mueller, Franziska and Davis, Micah and Bernard, Florian and Sotnychenko, Oleksandr and Verschoor, Mickeal and Otaduy, Miguel A and Casas, Dan and Theobalt, Christian},
  journal={ACM Transactions on Graphics (ToG)},
  volume={38},
  number={4},
  pages={1--13},
  year={2019},
  publisher={ACM New York, NY, USA}
}

@INPROCEEDINGS{reizenstein:2021,
  author={Reizenstein, Jeremy and Shapovalov, Roman and Henzler, Philipp and Sbordone, Luca and Labatut, Patrick and Novotny, David},
  booktitle={2021 IEEE/CVF International Conference on Computer Vision (ICCV)}, 
  title={Common Objects in 3D: Large-Scale Learning and Evaluation of Real-life 3D Category Reconstruction}, 
  year={2021},
  volume={},
  number={},
  pages={10881-10891},
  doi={10.1109/ICCV48922.2021.01072}}

@INPROCEEDINGS{Qu:2023,
  author={Qu, Wentian and Cui, Zhaopeng and Zhang, Yinda and Meng, Chenyu and Ma, Cuixia and Deng, Xiaoming and Wang, Hongan},
  booktitle={2023 IEEE/CVF International Conference on Computer Vision (ICCV)}, 
  title={Novel-view Synthesis and Pose Estimation for Hand-Object Interaction from Sparse Views}, 
  year={2023},
  volume={},
  number={},
  pages={15054-15065},
  doi={10.1109/ICCV51070.2023.01386}}

@article{prakash2023learning,
  title={Learning hand-held object reconstruction from in-the-wild videos},
  author={Prakash, Aditya and Chang, Matthew and Jin, Matthew and Gupta, Saurabh},
  journal={arXiv preprint},
  year={2023}
}

@article{wu2024reconstructing,
  title={Reconstructing hand-held objects in 3d from images and videos},
  author={Wu, Jane and Pavlakos, Georgios and Gkioxari, Georgia and Malik, Jitendra},
  journal={arXiv preprint arXiv:2404.06507},
  year={2024}
}

@inproceedings{jiang2024hand,
  title={In-hand 3D object reconstruction from a monocular RGB video},
  author={Jiang, Shijian and Ye, Qi and Xie, Rengan and Huo, Yuchi and Li, Xiang and Zhou, Yang and Chen, Jiming},
  booktitle={Proceedings of the AAAI Conference on Artificial Intelligence},
  volume={38},
  number={3},
  pages={2525--2533},
  year={2024}
}

@INPROCEEDINGS {xia:2024,
author = { Xia, Hongchi and Fu, Yang and Liu, Sifei and Wang, Xiaolong },
booktitle = { 2024 IEEE/CVF Conference on Computer Vision and Pattern Recognition (CVPR) },
title = {{ RGBD Objects in the Wild: Scaling Real-World 3D Object Learning from RGB-D Videos }},
year = {2024},
volume = {},
ISSN = {},
pages = {22378-22389},
doi = {10.1109/CVPR52733.2024.02112},
url = {https://doi.ieeecomputersociety.org/10.1109/CVPR52733.2024.02112},
publisher = {IEEE Computer Society},
address = {Los Alamitos, CA, USA},
month =Jun}

\end{document}